%% file: main.tex
\documentclass[letterpaper, 10pt, conference]{ieeeconf}      

\IEEEoverridecommandlockouts 

\usepackage{times}
\usepackage{tabularx}
\usepackage{subcaption}
\usepackage{multirow,multicol, array}
\usepackage[font={small}]{caption}   
\usepackage{graphicx,dblfloatfix}
\usepackage{wrapfig}
\usepackage{siunitx}
\usepackage{soul}

\let\labelindent\relax
\usepackage{enumitem}

\usepackage{amsmath,amsthm,amssymb,amsfonts}
\usepackage[titlenumbered,ruled]{algorithm2e}
\usepackage{textcomp}
\usepackage{acronym}
\usepackage{balance}
\usepackage{mdwmath}
\usepackage{bm}
\usepackage{mdwtab}
\usepackage{array}
\usepackage{eqparbox}
\usepackage{cite}
\usepackage{verbatim}
\usepackage{amssymb}

\usepackage{color}
\usepackage{xcolor}
\usepackage{psfrag}
\usepackage{epsfig}
\usepackage{url}

\usepackage{epstopdf}
\usepackage{booktabs}
\usepackage{blindtext}

\usepackage{physics}

\usepackage[colorinlistoftodos,prependcaption,textsize=tiny]{todonotes}

\renewcommand{\emph}{\textit}

\newtheorem*{lemma*}{Lemma}

\newtheorem*{problem*}{Problem}

\makeatletter
\newcommand\fs@spaceruled{\def\@fs@cfont{\bfseries}\let\@fs@capt\floatc@ruled
    \def\@fs@pre{\vspace{5\baselineskip}\hrule height.8pt depth0pt \kern2pt}%
    \def\@fs@post{\kern2pt\hrule\relax}%
    \def\@fs@mid{\kern2pt\hrule\kern2pt}%
    \let\@fs@iftopcapt\iftrue}
\def\maketag@@@#1{\hbox{\m@th\normalfont\normalsize#1}}
\makeatother

\IEEEoverridecommandlockouts
\graphicspath{{images/}}

\title{
Evaluation of Legged Robot Landing Capability 
Under \\Aggressive Linear and Angular Velocities 
}
	
\author{Keran Ye and Konstantinos Karydis
	\thanks{The authors are with the Dept. of Electrical and Computer Engineering, University of California, Riverside. 
	Email: {\{kye007, karydis\}@ucr.edu}.
	}
	\thanks{
	We gratefully acknowledge the support of NSF under grant \#CMMI-2046270. Any opinions, findings, and conclusions or recommendations expressed in this material are those of the authors and do not necessarily reflect the views of the National Science Foundation.}
}
	
\begin{document}

\maketitle
\thispagestyle{empty}
\pagestyle{empty}


\begin{abstract}
This paper proposes a method to evaluate the capability of aggressive legged robot landing under significant touchdown linear and angular velocities upon impact. Our approach builds upon the Planar Inverted Pendulum with Flywheel (PIPF) model and introduces a landing framework for the first stance step on a non-dimensional basis. We develop a nonlinear framework with iterative constrained trajectory optimization to stabilize the first stance step prior to N-step Capturability analysis. Performance maps across many different initial conditions reveal approximately linear boundaries as well as the effect of inertia, body incidence angle and leg attacking angle on the boundary shape. Our method also yields the engineering insight that body inertia affects the performance map the most, hence its optimization can be prioritized when the target is to improve robot landing efficacy.
\end{abstract}

\section{Introduction}
Legged robots have been developed for achieving high agility~\cite{ijspeert2014biorobotics,hwangbo2019learning,gehringoptimization,reher2021dynamic,haldane2016robotic} in locomotion such as fast running~\cite{park2017high,sreenath2013embedding}, quick stepping~\cite{siekmann2021blind,bledt2020extracting}, back flipping~\cite{katz2019mini,chignoli2021humanoid}, aggressive jumping~\cite{rudin2021cat,nguyen2019optimized,yim2018precision}, and large-perturbation recovery~\cite{stephens2010push,bledt2020extracting,lee2019robust}. 
The increased need for maneuverability and reactivity when operating in dynamic environments requires the robots to perform challenging locomotion tasks which can expose them to substantial energy and momentum that 
may in fact inhibit locomotion or even cause structural damage~\cite{kolvenbach2019towards}. 

To protect legged robots from such situations, emergency stop under safety constraints~\cite{stephens2010push,lee2019robust,nguyen2019optimized} has been used in practice, especially for jumping and recovery from perturbations. 
Several landing (following a jump) controllers have been proposed to either harness robot impedance~\cite{gehring2016practice,grimminger2020open}, embed system dynamics into optimization problems~\cite{xiong2018bipedal,chignoli2021humanoid,nguyen2019optimized}, or adopt trained policies in a reinforcement learning paradigm~\cite{rudin2021cat}.
Push-recovery strategies also integrate optimization-based~\cite{stephens2010push,bledt2020extracting} and learning-based methods~\cite{rebula2007learning,lee2019robust}. Meanwhile, planning of multiple foot placements like N-step Capturability~\cite{koolen2012capturability} has been investigated as a means to account for large perturbations unable to recover from with no-stepping balance controls~\cite{stephens2010dynamic,pratt2006capture}. 
To our knowledge, existing approaches either focus on specific direction of motion, like horizontal~\cite{stephens2010push,koolen2012capturability,rebula2007learning} or vertical~\cite{xiong2018bipedal,grimminger2020open}, or consider real-robot implementation but with mild-to-moderate initial conditions~\cite{raibert1984experiments,lee2019robust,bledt2020extracting,stephens2010dynamic}.

On the other hand, there has been limited attention to legged robot landing under both aggressive linear and angular momentum. 
The reflex-recovery strategy in~\cite{abdallah2005biomechanically} deals with external force disturbance that causes abrupt angular momentum changing rate; however, the magnitude of angular momentum may remain limited due to the short duration of the applied disturbance.
%
Biological evidence shows that some legged animals can deal with aggressive landing conditions elegantly and with little to no structural damage to their body~\cite{mckinley1983visual,bijma2016landing}. 
To further advance legged robots, it is important to evaluate the limits of a robot's landing capability first, before improving controllers and hardware toward their maximum potential and enhancing the robot's locomotion.

\input{Fig_inputs/fig_input_PIPF_model}

Methods for evaluation of landing capabilities should be able to generalize, considering that legged robots may vary widely in their detailed models (i.e. anchors~\cite{full1999templates}), and testing a proposed method on each and every anchor model would be too laborious. 
In contrast, templates like LIP~\cite{kajita1991study} and SLIP~\cite{schwind1999spring} are simpler yet shown sufficiently general to capture fundamental legged locomotion principles~\cite{sharbafi2017bioinspired}. 
The target template employed in this work is the Planar Inverted Pendulum with Flywheel (PIPF)~\cite{pratt2006capture} (Fig.~\ref{fig:model_PIPF}). 
We also adopt a non-dimensional analysis of critical variables and implement our method on a non-dimensional basis~\cite{pratt2006capture,huang2014simple,geyer2005spring} to account for a range of legged robot scales.

This paper focuses on legged robot fast landing stabilization from aggressive conditions (i.e. large body angular velocity and touch-down linear velocity) or at least relaxing them within the first step to milder ones (i.e. only horizontal velocity remaining) that are suitable for existing methods like~\cite{koolen2012capturability, sugihara20213d}.
%
%
%
\emph{The main technical result of this paper is a nonlinear optimization-based landing framework for the first stance step}. Our proposed framework contributes to:
1) first stance step stabilization for pitch and vertical motion with large horizontal velocity, 
2) iterative constrained landing trajectory optimization with unknown control horizon, 
3) vertical stabilization overall horizon bounds and feasibility conditions, 
4) performance maps over aggressive initial conditions for factor effect analysis, and 
5) engineering insights on robot design and control to improve landing.




\section{System Modeling}\label{sec:Model_Background}
Templates in sagittal plane serve as simplified representations to describe the underlying legged locomotion principles~\cite{full1999templates}. Among them, the Spring-loaded Inverted Pendulum (SLIP)~\cite{schwind1999spring} is mostly acknowledged for legged running with a massless elastic leg and a point-mass body, while its variants (e.g.,~\cite{seipel2007simple,huang2014simple,poulakakis2009spring}) help describe more complex behaviors for periodic gaits.
The Linear Inverted Pendulum (LIP)~\cite{kajita1991study} is a widely accepted template for legged walking. It consists of a massless linear leg and a point-mass body that is usually constrained in the horizontal plane to yield linear dynamics. The Planar Inverted Pendulum (PIP)~\cite{pratt2006capture} serves as a less constrained alternative to LIP allowing the body to move freely in the sagittal plane. Body pitch dynamics can be introduced into LIP and PIP through an inertial flywheel replacing the point-mass body~\cite{pratt2006capture}.

\subsection{Planar Inverted Pendulum with Flywheel (PIPF)}\label{subsec:PIPF}
PIP with Inertial Flywheel (PIPF, see Fig.~\ref{fig:model_PIPF})~\cite{pratt2006capture} covers the translational motions in both horizontal and gravitational directions as well as rotational motion around the transverse axis, and thus, is suitable as the target model in this paper. 
Here we revisit PIPF's Euler-Lagrange Equations of Motion (EoMs). The dynamics can be formed as
\begin{align}
    &\mathbf{M} \ddot{\mathbf{Q}}+\mathbf{b}+\mathbf{g} = \mathbf{F}(\mathbf{U})\enspace, \label{equ:PIPF_EoM}
\end{align}
\vspace{-22pt}
\begin{align}
\mathbf{M}\hspace{-1pt}&=
\hspace{-1pt}
\left[
\begin{array}{ccc} 
m & 0 & 0\\ 
0 & m{r}^2 & m{r}^2\\ 
0 & 0 & I 
\end{array}
\right]
\hspace{-1pt}, 
\quad
\mathbf{b}\hspace{-1pt}=
\hspace{-1pt}
\left[
\begin{array}{c} 
-mr{(\dot{\beta }+\dot{\gamma })}^2\\
2mr\dot{r}(\dot{\beta }+\dot{\gamma })\\
0 
\end{array}
\right], \nonumber \\ 
\mathbf{g}\hspace{-1pt}&=\hspace{-1pt}
\left[
\begin{array}{c} 
gm\sin(\beta +\gamma )\\ 
g(m\cos(\beta )\cos(\gamma )r-m\sin(\beta )\sin(\gamma )r)\\ 
0 
\end{array}
\right], \nonumber \\ 
\mathbf{F}&(\mathbf{U})\hspace{-1pt}=\hspace{-1pt}
\left[\quad
F, \quad
\tau, \quad
-\tau \quad 
\right]^{T}\enspace, \label{equ:PIPF_EoM_terms}
\vspace{-10pt}
\end{align}
where $\mathbf{M}$ is the generalized mass matrix, $\mathbf{b}$ the Coriolis-centrifugal vector, $\mathbf{g}$ the gravitational vector, and $\mathbf{F}$ the generalized force vector.
With reference to Fig.~\ref{fig:model_PIPF}, 
the configuration space is $\mathbf{Q}:=\left[r,\beta,\gamma \right]^T$ where $r$ is the leg length, $\beta$ the hip joint angle, and $\gamma$ the pitch angle of the flywheel. 
The input vector $\mathbf{U}:=\left[F,\tau \right]^T$ includes the prismatic leg force $F$ and the hip joint torque $\tau$. 
Scalar $v$ is the magnitude of the linear velocity $\mathbf{v}$, of which $v_x$ and $v_z$ are the horizontal and vertical components. $\omega = \dot{\gamma}$ is the body angular velocity. 
$\theta$ is the tilting-down angle of $\mathbf{v}$ with respect to the horizontal plane. $\theta_0$ is the body incidence angle (discussed in Section~\ref{subsec:bound_map_factors}).

Parameters $m$, $I$, $g$ and $r_0$ are mass, moment of inertia, gravity constant and initial leg length, respectively.
The Cartesian state vector of the flywheel center then is 
\begin{align}
\left[
\begin{array}{c} 
x \\
z \\
\dot{x} \\
\dot{z}
\end{array}
\right]
&=
\hspace{-1pt}
\left[
\begin{array}{ccc} 
x_f - r \cos(\alpha) \\ 
z_f + r \sin(\alpha) \\ 
\dot{x}_f - ( \dot{r} \cos(\alpha) - r \dot{\alpha} \sin(\alpha) ) \\
\dot{z}_f + ( \dot{r} \sin(\alpha) + r \dot{\alpha} \cos(\alpha) )
\end{array}
\right] \enspace,\label{equ:Cartesian_states}
\hspace{-1pt}
\end{align}
where $\mathbf{X}_c:= \left[x, z, \dot{x}, \dot{z} \right]^T$ includes the horizontal and vertical positions and velocities of the flywheel center in the inertial frame, $\mathbf{X}_f:= \left[x_f, z_f, \dot{x}_f, \dot{z}_f \right]^T$ the same Cartesian state vector for the leg's point foot, and $\alpha := \beta + \gamma$ is the attacking angle at the point foot. Note that $\{ \dot{x}, \dot{z} \}$ and $\{ v_x, v_z \}$ are used interchangeably in this paper.

\subsection{Non-dimensional Analysis}\label{subsec:Dimless_analysis}
Non-dimensional analysis over state variables and model parameters is a more generalized way to investigate system behaviors. Our method is implemented on a non-dimensional basis to account for different parameter scales. Similar to~\cite{pratt2006capture,huang2014simple}, we non-dimensionalize length, linear velocity, angular velocity, force, torque, inertia, and angle as
\hspace{0pt}
\begin{align}
\Tilde{r} &= \frac{r}{r_0}, \:
\Tilde{v} = \frac{v}{\sqrt{g r_0}}, \:
\Tilde{\omega} = \frac{\omega}{\sqrt{g / r_0}}, \:  
\Tilde{F} = \frac{F}{m g}, \:
\Tilde{\tau} = \frac{\tau}{m g r_0}, \nonumber \\
\Tilde{I} &= \frac{I}{m r_0^2}, \: 
\Tilde{\alpha} = \alpha, \:
\Tilde{\beta} = \beta, \:
\Tilde{\gamma} = \gamma, \:
\Tilde{\theta} = \theta\enspace.
\hspace{0pt} \label{equ:dimensionless_analysis}
\vspace{-30pt}
\end{align}
Note that angles $\alpha$, $\beta$, $\gamma$, and $\theta$ can be treated as non-dimensional quantities by themselves. The time constant $T_C := \sqrt{r_0 / g}$ will be used to define the horizon of our trajectory optimization problem discussed next.


\section{Nonlinear Optimization-Based Landing \\for First Stance Step}\label{sec:Landing_Evaluation}



\subsection{Overview}\label{subsec:Workflow_overview}
Here we present the main technical result of the paper: a nonlinear optimization-based framework for the first stance step stabilization. 
We aim to enforce PIPF into a static posture in finite steps, under the working conditions that 1) the overall kinematic energy is too massive to dissipate in a single step, and 2) the rotational momentum is substantial and risks causing body flip forward. 
The work of N-step capturability~\cite{koolen2012capturability} has laid out a solid foundation on stabilizing the LIP model in the horizontal plane. 
Therefore, we focus on the first stance step upon landing and seek to stabilize the PIPF's rotational and vertical motions in the presence of horizontal motion, as a way to relax aggressive conditions (large angular velocity and touch-down linear velocity) to a milder one (horizontal linear velocity only) that is reasonable to be handled with the N-step capturability method.

\begin{figure}[!t]
	\vspace{0pt}
	\centering
\includegraphics[trim={0cm 0cm 0cm 0cm},clip,width=\linewidth]{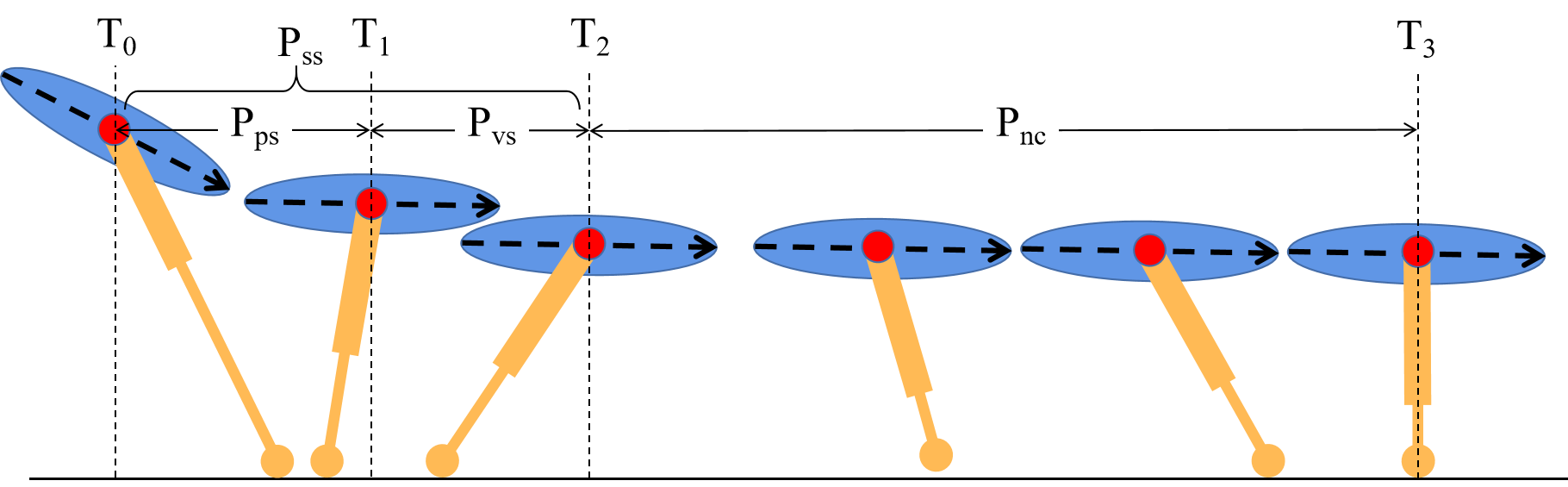}
	\vspace{-10pt}
	\caption{\textbf{Landing Phases for PIPF.} 
	The first stance step $P_{ss}$ ($T_0$-$T_2$) comprises the pitch stabilization $P_{ps}$ ($T_0$-$T_1$, overall horizon length $T_{ps}$) and vertical stabilization phases $P_{vs}$ ($T_1$-$T_2$, overall horizon length $T_{vs}$).
	Phase $P_{nc}$ ($T_2$-$T_3$) represents the N-step capturability (N = 1 here).
	The paper focuses on the first stance step.
	}
	\label{fig:workflow_landing_sequence}
	\vspace{-15pt}
\end{figure}

Figure~\ref{fig:workflow_landing_sequence} demonstrates landing phases---the first stance step $P_{ss}$ is where this paper focuses on. 
Correspondingly, we propose a nonlinear optimization-based framework for the first stance step stabilization. 
Figure~\ref{fig:evaluation_workflow} summarizes the workflow with three major components being: 
\begin{enumerate}
    \item \textbf{Pitch Stabilization} aiming to stabilize both pitch and vertical motions in the phase $P_{ps}$ before $T_1$ (Section~\ref{subsec:pitch_stabilization}). Pitch motion is prioritized to avoid body overturning that could be catastrophic to onboard sensing from an engineering perspective. 
    \item \textbf{Vertical Stabilization Feasibility Check} investigating the practicability of vertical stabilization at $T_1$ and predict its possible horizon $T_{vs}$ (Section~\ref{subsec:vertical_feasibility_check}).
    \item \textbf{Vertical Stabilization} stabilizing remaining vertical motion in the phase $P_{vs}$ before $T_2$  (Section~\ref{subsec:vertical_stabilization}).
\end{enumerate}

Table~\ref{table:variable_summary} summarizes the important variables. 
%


\input{Tables/states_summary}

\subsection{Pitch Stabilization}\label{subsec:pitch_stabilization}

\begin{figure}[!t]
	\vspace{0pt}
	\centering
\includegraphics[trim={0cm 0cm 0cm 0cm},clip,width=\linewidth]{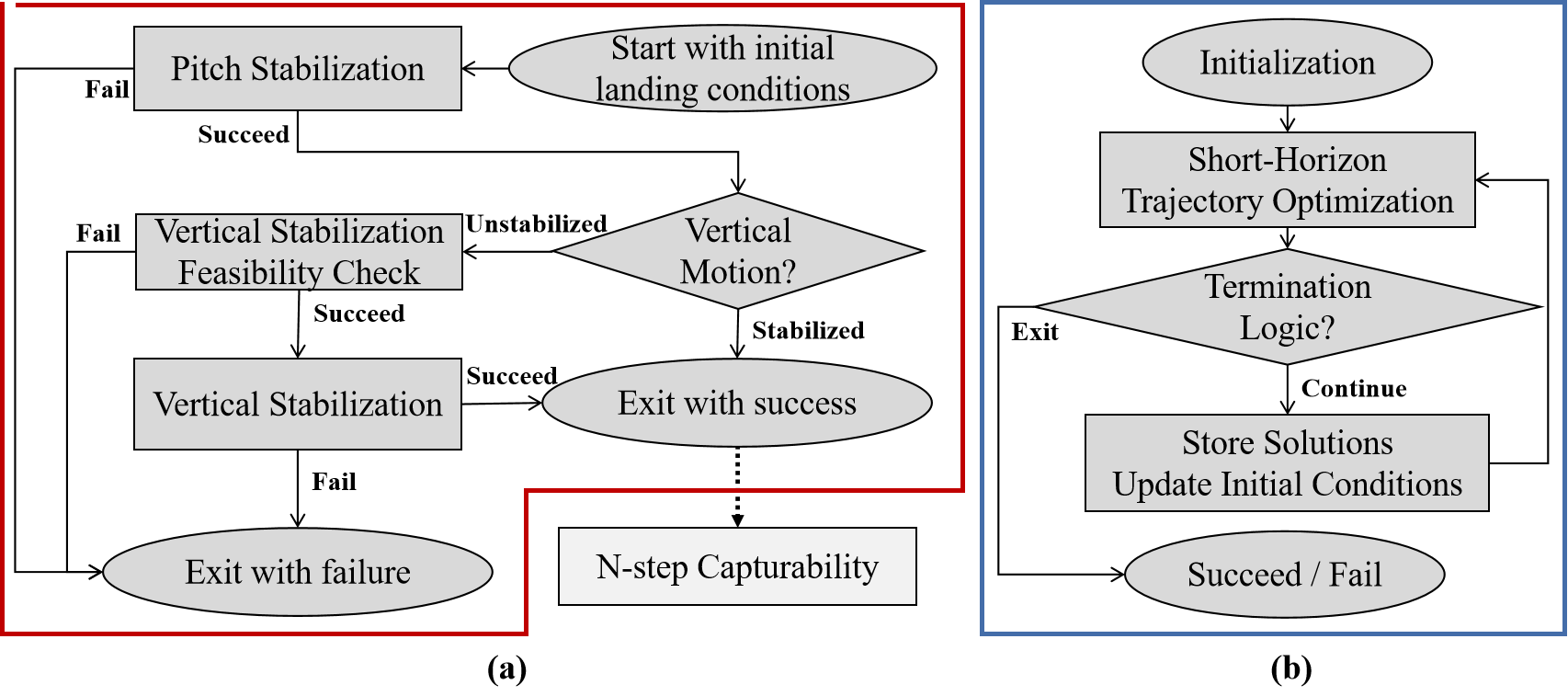}
	\vspace{-15pt}
	\caption{\textbf{First Stance Step Stabilization Workflow.} (a) The workflow of the first stance step stabilization. (b) The iterative nonlinear trajectory optimization for the pitch/vertical stabilization.}
	\label{fig:evaluation_workflow}
	\vspace{-12pt}
\end{figure}

We adopt constrained nonlinear trajectory optimization for pitch stabilization. 
No assumption is made on the nominal state trajectories
\footnote{~We do not assume any prior knowledge for the landing trajectories and landing height so that the optimizer could search for solutions within wider state regions instead of the neighborhood of the nominal trajectories. 
Only the desired landing posture is known as no rotation and vertical motion.
}
nor the control horizon $T_{ps}$,
so we seek to iterate the optimization over a small control horizon with updated initial values and termination conditions (Fig.~\ref{fig:evaluation_workflow}b).
Below is the optimization setup within one iteration.

First, the small control horizon length $T_h$ is determined for the current iteration.
Considering the horizontal motion is not the major concern for pitch stabilization, we set $\Tilde{v}_{x0}$, the horizontal component of $\Tilde{v}_{0}$, as a rough guess of the average non-dimensional horizontal velocity. 
Then, we define $T_h$ for each iteration during $P_{ps}$ via
\vspace{-5pt}
\begin{align}
T_h |_{P_{ps}} = 
\eta \frac{1}{\Tilde{v}_{x0}} T_C\enspace. \label{equ:control_horizon_pitch}
\vspace{-8pt}
\end{align}
$\frac{1}{\Tilde{v}_{x0}}$ represents the non-dimensional time to traverse a unit length at speed $\Tilde{v}_{x0}$. $T_C$ is the time constant and $\eta$ is the percentage taken for each iteration.
The control horizon is $T_h = T_s p, p \in \mathbb{N}$ and $T_s$ is the time step length. 

Given the initial time $t_0$, the continuous-time trajectory optimization problem is formulated over $\left[ t_0, t_0 + T_h \right]$ as
\vspace{-6pt}
\begin{align}
\vspace{-10pt}
\begin{array}{l}
\min _{\mathbf{X}, \mathbf{U}} J(\mathbf{X})\\
\text {subject to } \dot{\mathbf{X}} = \mathbf{f}(\mathbf{X},\mathbf{U}), \:
\mathbf{C}_{ineq}(\mathbf{X},\mathbf{U}) \leq \mathbf{0}
\vspace{0pt}
\end{array}\label{equ:rigid_nlmpc}
\end{align}
where states $\mathbf{X}=\left[ \mathbf{Q}, \dot{\mathbf{Q}} \right]^T$, configuration states $\mathbf{Q}$ and inputs $\mathbf{U}$ are defined in Section~\ref{subsec:PIPF}, 
and dynamic constraints $\dot{\mathbf{X}} = \mathbf{f}(\mathbf{X},\mathbf{U})$ based on EoMs \eqref{equ:PIPF_EoM}--\eqref{equ:PIPF_EoM_terms}:
\begin{align}
\mathbf{f}
=\left[
\begin{array}{c}
\dot{\mathbf{Q}} \\ 
\mathbf{M}^{-1}(\mathbf{F}(\mathbf{U}) - \mathbf{b} - \mathbf{g})
\end{array}
\right]\enspace. \label{equ:dynamic_constraint}
\end{align}

The discretized cost function $\mathbf{J} := \mathbf{J}_L + \mathbf{J}_M$ adopts the standard quadratic form of Lagrangian and Mayer terms as 
\begin{align}
\mathbf{J}_{L} &= 
k_1 \sum_{n=1}^{p-1} \xi_{\dot{z}, n} \sigma^2_{\Tilde{\dot{z}}, n} +
k_2 \sum_{n=1}^{p-1} \xi_{\gamma, n} \sigma^2_{\Tilde{\gamma}, n} +
k_3 \sum_{n=1}^{p-1} \xi_{\dot{\gamma}, n} \sigma^2_{\Tilde{\dot{\gamma}}, n} \nonumber \\
\sigma^2_{\Tilde{\dot{z}}, n} &= 
(\Tilde{\dot{z}}^{d}-\Tilde{\dot{z}}_{n})^2,
\sigma^2_{\Tilde{\gamma}, n} = 
(\Tilde{\gamma}^{d}-\Tilde{\gamma}_{n})^2,
\sigma^2_{\Tilde{\dot{\gamma}}, n} = 
(\Tilde{\dot{\gamma}}^{d}-\Tilde{\dot{\gamma}}_{n})^2 \nonumber \\
\mathbf{J}_{M} &= 
k_1 \xi_{\dot{z}, p} \sigma^2_{\Tilde{\dot{z}}, p} +
k_2 \xi_{\gamma, p} \sigma^2_{\Tilde{\gamma}, p} +
k_3 \xi_{\dot{\gamma}, p} \sigma^2_{\Tilde{\dot{\gamma}}, p}
\label{equ:cost_function}
\hspace{-1pt}
\end{align}
where 
$\mathbf{J}_{L}$ is a weighted sum of the discounted average deviation along the control horizon and $\mathbf{J}_{M}$ applies only to the final instant. 
$\sigma^2_{\Tilde{\dot{z}}, n}$, $\sigma^2_{\Tilde{\gamma}, n}$, and $\sigma^2_{\Tilde{\dot{\gamma}}, n}$ are deviations of $\Tilde{\dot{z}}$, $\Tilde{\gamma}$, and $\Tilde{\dot{\gamma}}$ (computed by~\eqref{equ:Cartesian_states}--\eqref{equ:dimensionless_analysis}) about their desired final landing posture values
denoted with superscript $^d$ at step $n$. 
Discounts $\xi_{\dot{z}, n}$, $\xi_{\gamma, n}$, and $\xi_{\dot{\gamma}, n}$ adjust the attention level at each time step along the current horizon. 
Weights $k_1 < k_2, k_1 < k_3$ indicate that pitch states are the priority to stabilize. 

Note that 
discount sequences $\{ \xi_{\dot{z}, n} \}$, $\{ \xi_{\gamma, n} \}$ and $\{ \xi_{\dot{\gamma}, n} \}$ belong to 
either Uniform or Poisson distributions, so that each sequence sums up strictly or almost to 1 and thus accounts for discounted averaging on each non-dimensional variable. Uniform distribution suggests equal attention of the solver for every step, and Poisson distribution introduces a gradual change of attention on different steps. 
In practice, we adopt a reversed Poisson (with $\lambda = 1$) sequence such that the solver focuses more on the last few steps. 
Observations from various trials suggest that 1) Uniform distribution performs reliably on stabilizing variables with small initial deviation, like $\gamma$ and $\dot{\gamma}$ during $P_{vs}$, and 2) Poisson distribution is more appropriate for cases with large initial deviation, like the states in $P_{ps}$, which is probably due to more flexibility on early steps. 
Also note that dynamic constraints~\eqref{equ:dynamic_constraint} with~\eqref{equ:Cartesian_states} imply the coupled relationship between
the horizontal and vertical states. The optimization will keep this logic even though the states are distinguished in the objective function.

We require the model state trajectories to be properly constrained in the sense that 1) the body orientation allows no excess tilting toward overturning, 2) the hip joint does not rotate the leg over the body's transverse plane, and 3) actuation forces are limited. The major constraints $\mathbf{C}_{ineq}(\mathbf{X},\mathbf{U}) \leq \mathbf{0}$ include Geometric limits on the configuration space $\mathbf{Q}$ and actuation limits. More specifically,
%
%
%
\vspace{-2pt}
\begin{align}
    &\Tilde{\mathbf{Q}} \in \left[ \Tilde{\mathbf{Q}}_{min}, \Tilde{\mathbf{Q}}_{min} \right], \quad
    \Tilde{\mathbf{Q}}:=\left[\Tilde{r},\Tilde{\beta},\Tilde{\gamma} \right]^T\;,
    \label{equ:config_geo_constraints}
    \\
    &\Tilde{\mathbf{U}} \in \left[ \Tilde{\mathbf{U}}_{min}, \Tilde{\mathbf{U}}_{min} \right], \quad
    \Tilde{\mathbf{U}}:=\left[ \Tilde{F}, \Tilde{\tau} \right]^T\enspace.
    \label{equ:actuation_constraints}
\end{align}

\begin{figure}[!h]
	\vspace{-6pt}
\includegraphics[trim={1cm 0cm 1.5cm 0cm},clip,width=0.499\linewidth]{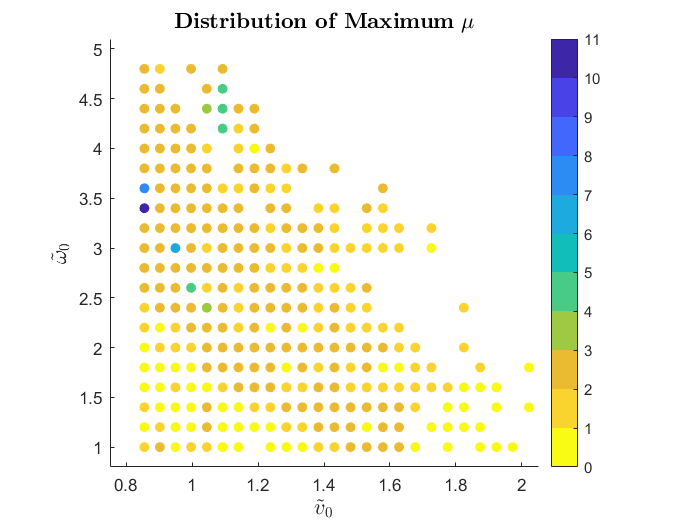}
\includegraphics[trim={1cm 0cm 1.5cm 0cm},clip,width=0.49\linewidth]{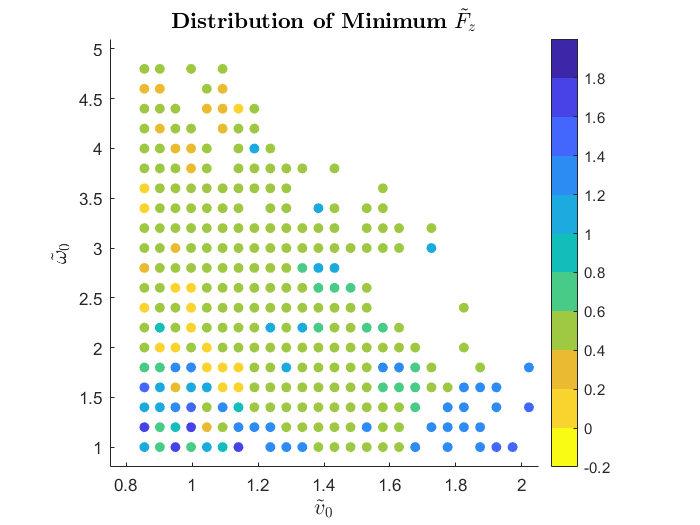}
	\vspace{-15pt}
	\caption{\textbf{Distributions of Factors about Ground Reaction Force (GRF).} Our optimization method excludes GRF-computed constraints, and the data collection is based on successful cases with initial conditions $\Tilde{\omega}_{0} \in [1,5]$, $\Tilde{v}_{x0} \in [0.8,2.0]$, and $\Tilde{v}_{z0} = -0.3$.  The left panel illustrates the maximum $\mu = | \frac{F_{fx}}{F_{fz}} |$ to denote the required friction coefficient, with most cases marked in warm tones. The right panel shows the minimum non-dimensional normal force $\Tilde{F}_{fz}$ at the foot, with no case marked in yellow as negative. }
	\label{fig:distribution_GRF_factor}
	\vspace{-6pt}
\end{figure}

Additional constraints may involve computing Ground Reaction Forces (GRF) 
to indicate the Coulomb friction limits and normal supporting force on foot. 
Nevertheless, as shown in Fig.~\ref{fig:distribution_GRF_factor}, we observe that optimization without GRF constraints still converges to a solution with positive $F_{fz}$. It also reveals that in most cases the required friction coefficient is less than 3, which is within practically achievable ranges for a rubber foot on the clean ground~\cite{gough1960friction}. Thus, we run the optimization without GRF constraints that can increase complexity by requiring additional Jacobian matrices inversion.

The pitch stabilization process decides to continue or exit every time the trajectory optimization ends.
We let the attacking angle $\alpha$ be feasible when $ 0 < \alpha_{min} < \alpha < \alpha_{max} < \pi$, so that the foot always points downward. 
We also expect the neutral position $\alpha = \pi/2$ to be reached or passed through when phase $P_{ps}$ ends.  
As a result, the termination logic summarizes that the pitch stabilization will
\begin{itemize}
    \item exit with success, if the pitch dynamics is stabilized while $\alpha \geq \pi/2$ and before $\alpha$ becomes infeasible;
    \item exit with failure, if the pitch dynamics is still not stabilized when $\alpha$ becomes infeasible;
    \item continue for the next iteration, otherwise.
\end{itemize}
For the next iteration, initial state values will be updated with configuration states $\mathbf{Q}$, time derivative $\dot{\mathbf{Q}}$ and time variable $t$ at the final step of the current optimization solution.

\subsection{Vertical Stabilization Feasibility Check }\label{subsec:vertical_feasibility_check}

When the pitch stabilization ends successfully and vertical motion is yet to be stopped, that is when
\begin{align}
    v_{x1} > 0, \quad v_{z1} < 0,\quad \alpha_{1} > \frac{\pi}{2},\quad \dot{\alpha}_{1} > 0\enspace, \label{equ:vertical_init_conditions}
\end{align}
the subsequent vertical stabilization process takes over but may encounter several ill conditions. These can include: 1) $v_z$ is still quite large, 2) $\alpha$ is already close to the upper boundary $\alpha_{max}$, and 3) $\dot{\alpha}$ is rather large to sweep over the remaining distance in the first stance step. 
The above situations may require adjustment over the optimization setup (e.g., the control horizon) for the vertical stabilization in $P_{vs}$ and even undermine its feasibility.

Based on the states at $T_1$ (see Table~\ref{table:variable_summary}),
we provide the lower and upper bound estimation $T^{lb}_{vs}$ and $T^{ub}_{vs}$ for the overall control horizon length $T_{vs}$ of the vertical stabilization as
\begin{align}
    T^{lb}_{vs} &= 
    \frac{-\Tilde{v}_{z1}}{ \left(\Tilde{F}_{max} \sin{\Tilde{\alpha}_{1}} - 1 \right) } T_C
    \nonumber \\
    T^{ub}_{vs} &= 
    \frac{-\Tilde{\dot{\alpha}}_1 + \sqrt{ \Tilde{\dot{\alpha}}^2_1 + 2 \Tilde{\ddot{\alpha}}^{lb}_{vs} ( \Tilde{\alpha}^{ub}_{vs} - \Tilde{\alpha}_1)}} {\Tilde{\ddot{\alpha}}^{lb}_{vs}} T_C
    \label{equ:Tvs_lower_upper_bound}
\end{align}
where $\Tilde{\ddot{\alpha}}^{lb}_{vs}$ and  $\Tilde{\alpha}^{ub}_{vs}$ are non-dimensional attacking angle acceleration lower bound ($\ddot{\alpha}^{lb}_{vs}$) and displacement upper bound ($\alpha^{ub}_{vs}$), respectively, in phase $P_{vs}$, defined as\footnote{~The derivation of~\eqref{equ:Tvs_lower_upper_bound} and~\eqref{equ:alpha_alphaddot_lower_upper_bound} is based on model dynamics and kinematic constraints, and can be found at https://bit.ly/3eNT753.}
\begin{align}
    \ddot{\alpha}^{lb}_{vs} &= 
    \frac{ -\cos{\alpha_{1}} } {\left( 1 + \Tilde{I} \right) } \frac{\Tilde{r}_1}{T_C^2}, 
    \quad
    \alpha^{ub}_{vs} = 
    \frac{\pi}{2} + \cos^{-1}{(\Tilde{F}^{-1}_{max})}
    \label{equ:alpha_alphaddot_lower_upper_bound}
\end{align}

With the lower and upper bounds in place, we define that vertical stabilization feasibility check exits with success if
\begin{align}
    T^{lb}_{vs} < T^{ub}_{vs},
    \quad 
    T^{lb}_{vs} > 0,
    \quad 
    T^{ub}_{vs} > 0\enspace.
    \label{equ:vertical_stabilization_feasibility_condition}
\end{align}

\subsection{Vertical Stabilization}\label{subsec:vertical_stabilization}

The vertical stabilization process also adopts the same iterative optimization setup as in pitch stabilization (Section~\ref{subsec:pitch_stabilization}), modulo some adjustments on the control horizon, cost function weights and the termination logic explained next.

The small control horizon length $T_h$ of vertical stabilization is 
\vspace{-2pt}
\begin{align}
T_h |_{P_{vs}} = 
\eta T^{lb}_{vs}\enspace, \label{equ:control_horizon_vertical}
\vspace{-2pt}
\end{align}
where $\eta$ is the percentage factor same as~\eqref{equ:control_horizon_pitch}, and $T^{lb}_{vs}$ the lower bound estimate of the overall horizon length $T_{vs}$ computed from~\eqref{equ:Tvs_lower_upper_bound}.

Weights in the cost function~\eqref{equ:cost_function} are modified to satisfy $k_1 > k_2, k_1 > k_3$ so that the vertical motion is the priority to stabilize. 

\input{Fig_inputs/fig_input_result_state_traj}

The termination logic is adjusted so that the vertical stabilization will
\begin{itemize}
    \item exit with success, if the vertical motion is stabilized before $\alpha$ becomes infeasible;
    \item exit with failure, if the vertical motion is still not stabilized when $\alpha$ becomes infeasible;
    \item continue for the next iteration, otherwise.
\end{itemize}

\section{Results and Discussion}\label{sec:Results}

\begin{figure}[!t]
	\vspace{-4pt}
	\centering
\includegraphics[trim={0.5cm 2.8cm 1.0cm 3cm},clip,width=0.90\linewidth]{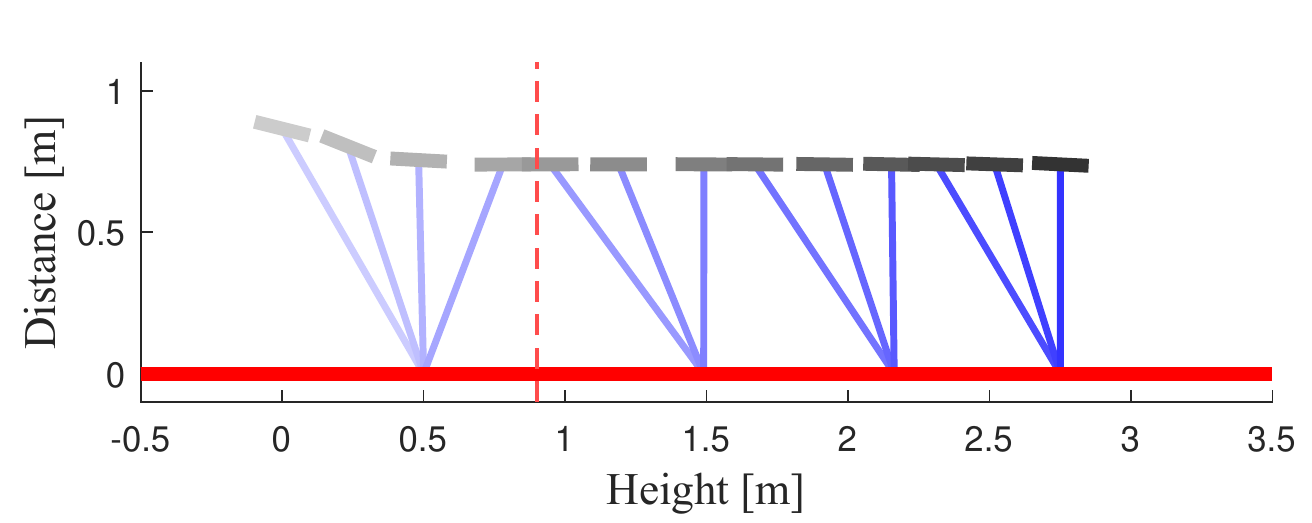}
	\vspace{0pt}
	\caption{\textbf{Preliminary Results of Landing Sequence.} Landing trajectories during the first stance step (left-hand side of red dashed line) and following $N=3$ steps (right-hand side of red dashed line) for PIPF model, with sample initial condition~\eqref{equ:prelim_test_init_cond}.
	Ground is shown as red line. (Figure best viewed in color.)}
	\label{fig:landing_sequence}
	\vspace{-10pt}
\end{figure}

\subsection{Selection of Initial Conditions}\label{subsec:select_w0_vx0_vz0}
The paper focuses on the aggressive conditions over $\Tilde{\omega}_0$, $\Tilde{v}_{x0}$ and $\Tilde{v}_{z0}$. 
For the angular velocity, we investigate the span $\Tilde{\omega}_{0} \in [1,5]$. 
For the vertical linear velocity, we study the range $\Tilde{v}_{z0} \in [-0.3,-0.5]$.
For the horizontal linear velocity, we determine the minimum $\Tilde{v}_{x0,min}$ based on the LIP model's Capture Point (according to~\cite[Eq. (10)]{pratt2006capture}) such that the Capture Point is beyond the leg reach at $T_0$ even for $v_{x0,min}$. Therefore we have
\vspace{-2pt}
\begin{align}
    & x_{capture} 
    = v_{x0,min} \sqrt{\frac{z_0}{g}}
    = \eta_{v_x} r_0 \cos{\alpha_0}, \,
    \eta_{v_x} > 1 \nonumber \\
    & v_{x0,min} 
    = \Tilde{v}_{x0,min} \sqrt{g r_0}, \quad
    z_0 = r_0 \sin{\alpha_0} \nonumber \\
    \Rightarrow 
    & \Tilde{v}_{x0,min} 
    = \eta_{v_x} \frac{\cos{\alpha_0}}{\sqrt{\sin{\alpha_0}}}
    \label{equ:capture_point}
\end{align}
where $\alpha_0$ is prefixed and $\eta_{v_x}$ is the amplifying factor. We set $\eta_{v_x} = 1.5$ and the gap $\Delta \Tilde{v}_{x0} = \Tilde{v}_{x0,max} - \Tilde{v}_{x0,min} = 1.2$, and compute the span $\Tilde{v}_{x0} \in [\Tilde{v}_{x0,min},\Tilde{v}_{x0,max}]$ as
\vspace{-2pt}
\begin{align}
    & \Tilde{v}_{x0} \in [0.80, 2.00], \quad \alpha_0 = 60^{\circ}\enspace; \nonumber \\
    & \Tilde{v}_{x0} \in [0.95, 2.15], \quad \alpha_0 = 55^{\circ}\enspace; \nonumber \\
    & \Tilde{v}_{x0} \in [1.10, 2.30], \quad \alpha_0 = 50^{\circ}\enspace.
    \label{equ:vx0_search_bounds}
\end{align}

\subsection{Preliminary Test}\label{subsec:prelim_test}

Throughout Section~\ref{sec:Results}, we maintain the consistency on the model and optimization parameters and limits as
\vspace{-2pt}
\begin{align}
    &m =  80\, [\text{kg}], \,
    r_0 = 1\, [\text{m}] , \,
    g = 9.8\, [\text{m}/\text{s}^2] \nonumber\\
    &\Tilde{r} \in [0.4,1], \quad
    \Tilde{\beta} \in [0,\pi], \quad
    \Tilde{\gamma} \in [-\pi/2,\pi/2], \nonumber \\
    &\Tilde{F} \in [0,2], \quad
    \Tilde{\tau} \in [-1,1]
    \label{equ:parameter_values}
    \vspace{-5pt}
\end{align}
where $m$, $r_0$, and $g$ are selected based on~\cite{geyer2005spring}. We also assume the initial pitch angle $\gamma_0 = \theta_0$, with incidence angle $\theta_0$ computed from selected $\Tilde{v}_{x0}$ and $\Tilde{v}_{z0}$.
For the cost function, we adopt the reversed Poisson ($\lambda = 1$) for all the discounts except $\xi_{\gamma, n}$ and $\xi_{\dot{\gamma}, n}$ during $P_{vs}$;
we also carefully select the cost weights and desired values in~\eqref{equ:cost_function} as
\begin{align}
    & \text{During } P_{ps}: 
    k_1 \in \{ 10^0, 10^1, 10^2, 10^3 \}, k_2 = 10^4, k_3 = 10^5 \nonumber \\
    & \text{During } P_{vs}: 
    k_1 \in \{ 10^6, 10^7, 10^8 \}, k_2 = 10^3, k_3 = 10^3 \nonumber \\
    & \Tilde{\dot{z}}^{d} = 0.01, \quad \Tilde{\gamma}^{d} = 0, \quad  \Tilde{\dot{\gamma}}^{d} = 0
    \label{equ:weight_target_values}
    \vspace{-5pt}
\end{align}
where we test a case with different scales for the vertical weight $k_1$ to rule out bad selections.
We use a small positive number for $\Tilde{\dot{z}}^{d}$ to better neutralize the vertical velocity.
Other parameters like initial attacking angle $\alpha_0$ and non-dimensional inertia $\Tilde{I}$ are tunable and investigated in Sections~\ref{subsec:select_w0_vx0_vz0} and~\ref{subsec:bound_map_factors}, respectively.

We show results from a set of aggressive initial conditions:
\vspace{-12pt}
\begin{align}
    &\Tilde{\omega}_0 = 3, \,
    \Tilde{v}_{x0} = 1.2 , \,
    \Tilde{v}_{v0} = -0.3, \,
    \Tilde{I} = 0.04, \,
    \alpha_{0} = 60^{\circ}. 
    \label{equ:prelim_test_init_cond}
    \vspace{-5pt}
\end{align}
Figure~\ref{fig:state_traj} depicts state trajectories in terms of different motion aspects during the first stance step. We observe that 1) the pitch and vertical motion are stabilized, 2) the horizontal velocity is kept close to the initial level, and 3) the body posture is never at risk of overturning.
Figure~\ref{fig:landing_sequence} illustrates a complete landing procedure composed of the first stance step and next N steps from LIP model's N-step capturability~\cite{koolen2012capturability}. 

\begin{figure*}[!t]
\vspace{-15pt}
\includegraphics[trim={19.5cm 6cm 20.5cm 0cm},clip,width=0.28\linewidth]{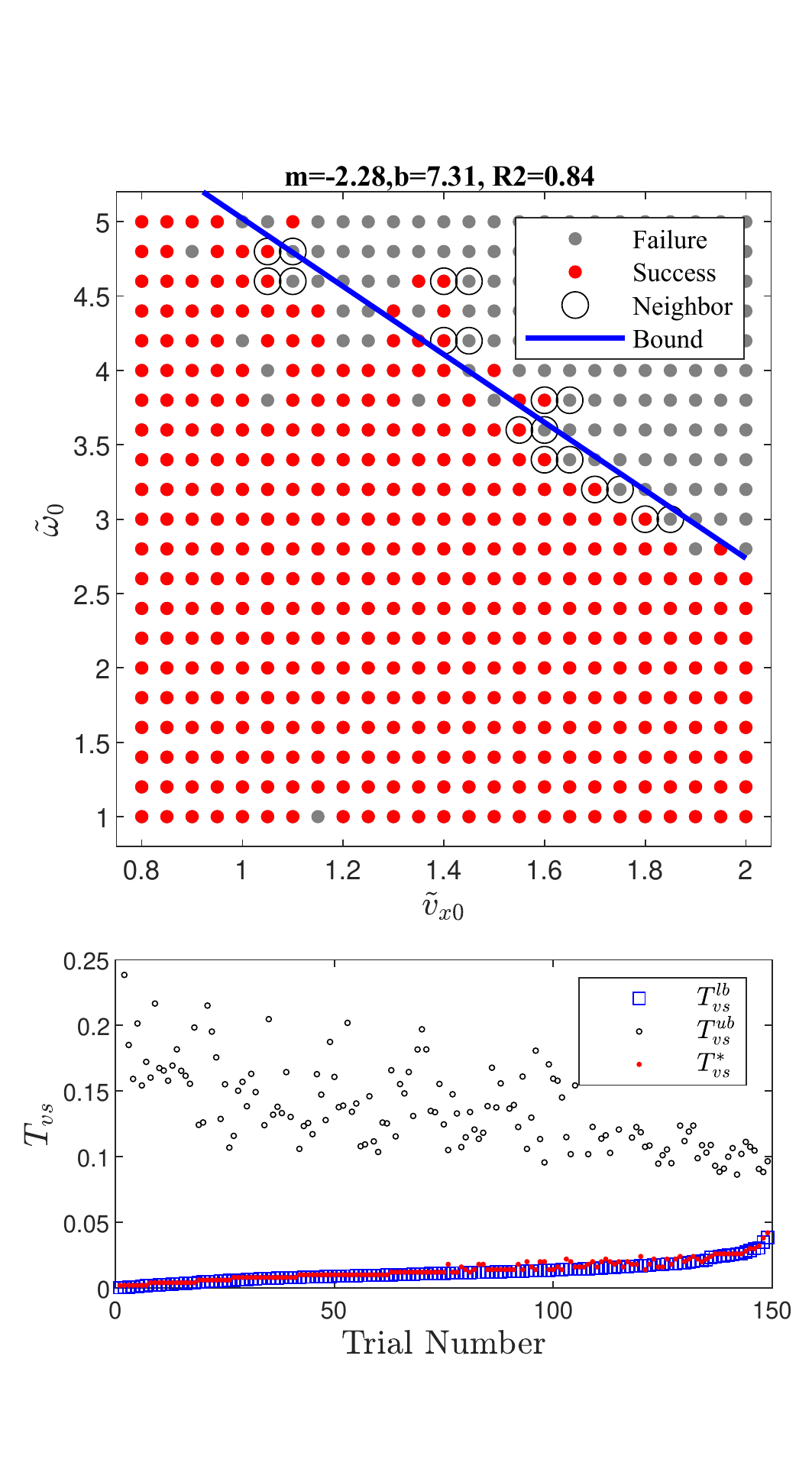}
\includegraphics[trim={1.7cm 0.3cm 2cm 0cm},clip,width=0.7\linewidth]{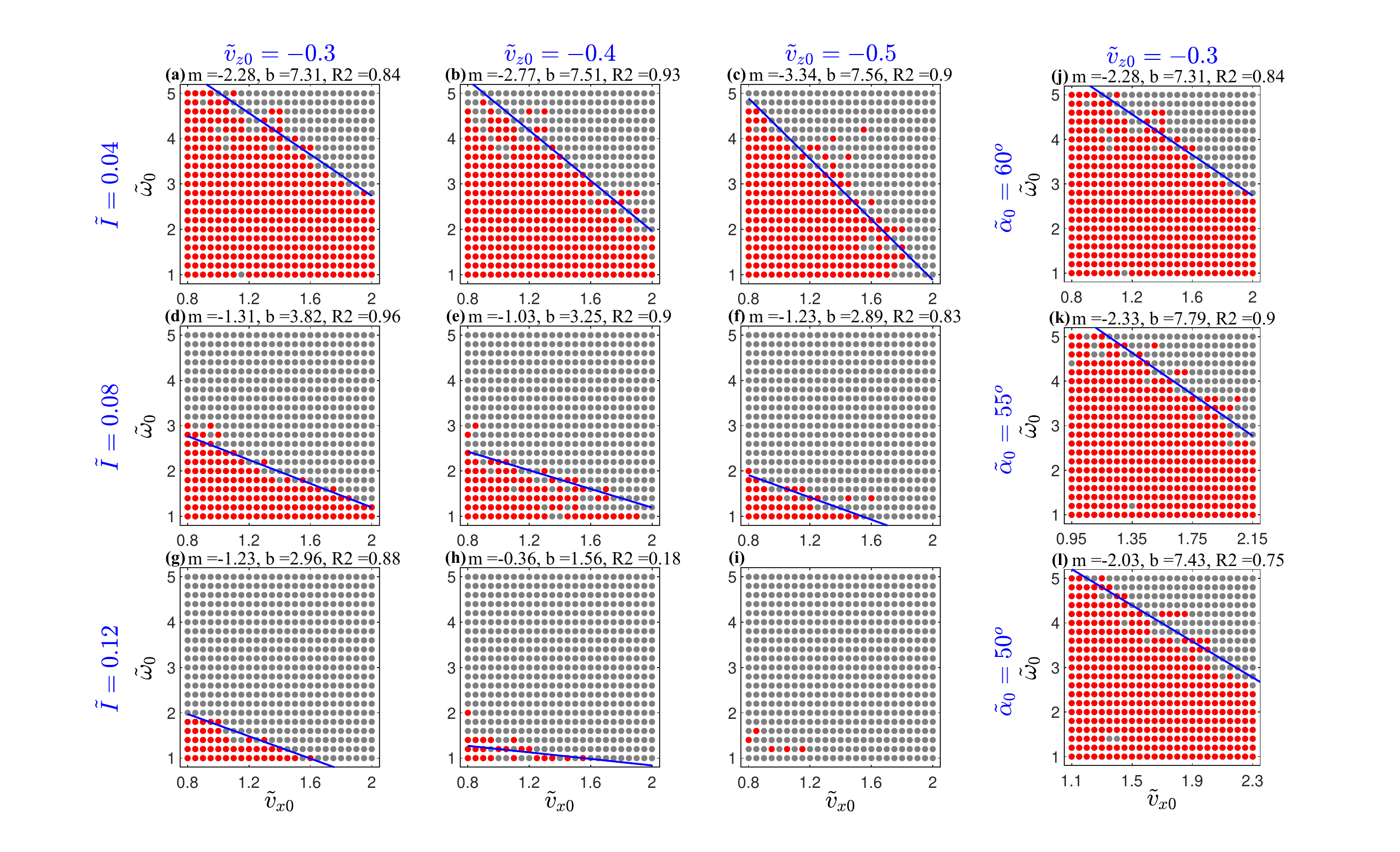}
\vspace{-12pt}
\caption{\textbf{Performance Map Analysis. }
\textbf{Left column:} 
A performance map (top) and $T_{vs}$ graph (bottom) over initial conditions~\eqref{equ:bound_map_init_cond}. The performance map includes tested cases, 
boundary neighbor cases (circled in black), and the approximated boundary line. 
The $T_{vs}$ graph is based on the lower bound $T^{lb}_{vs}$ (blue), upper bound $T^{ub}_{vs}$ (black), and optimized duration $T^{*}_{vs}$ (red) of the successful cases from the above performance map, with down-sampling and sorting by $T^{lb}_{vs}$ for better visualization.
\textbf{Middle 3 columns:} 
Panels (a)--(i) present 3x3 performance maps of initial conditions $\Tilde{\omega}_{0} \in [1,5]$ and $\Tilde{v}_{x0} \in [0.8,2.0]$: for each column, $\Tilde{v}_{z0} = -0.3, -0.4, -0.5$; for each row, $\Tilde{I} = 0.04, 0.08, 0.12$.
\textbf{Right column:} 
Panels (j)--(l) show the performance maps of $\Tilde{\omega}_{0} \in [1,5]$, $\Tilde{v}_{x0} \in [0.8,2.0]$, $\Tilde{I} = 0.04$, and $\Tilde{v}_{z0} = -0.3$. $\Tilde{I} = 0.04, 0.08, 0.12$ for (j), (k), and (l) respectively. 
\textbf{Note:}
In all performance maps, successful cases are marked in red and failing cases in grey. A linear bound (blue) is fitted for the data in each map, denoted with its slope $m$, intercept $b$, and variance $R^2$ rounded to two decimal places. Note that panel (i) has no boundary fitted due to insufficient data.
}
\label{fig:bound_map_collect}
\vspace{-18pt}
\end{figure*}

\subsection{Performance Map over Initial Conditions}\label{subsec:bound_map}

Here we investigate a performance map of various initial conditions, with parameter values as~\eqref{equ:parameter_values} and~\eqref{equ:weight_target_values}.
Like in Section~\ref{subsec:prelim_test}, we also apply $\Tilde{I}_{0} = 0.04$ and $\alpha_{0} = 60^{\circ}$. 
With~\eqref{equ:vx0_search_bounds}, we determine the search area as 
\begin{align}
     \Tilde{\omega}_{0} \in [1,5], \quad 
     \Tilde{v}_{x0} \in [0.8,2.0], \quad 
     \Tilde{v}_{z0} = -0.3\enspace.
     \label{equ:bound_map_init_cond}
\end{align}
\vspace{-1pt}
The left column of Fig.~\ref{fig:bound_map_collect} shows results of our first stance step landing for the above setup. 
The top panel depicts the map of different initial condition setups assessed as either success or failure. 
A linear boundary is approximated at the right-upper corner with linear regression on the data in the neighborhood of the boundary. 
In the bottom panel we compare the $P_{vs}$ duration bounds $T^{lb}_{vs}$ and $T^{ub}_{vs}$, and the optimized duration $T^{*}_{vs}$ across all tests in the same map. 
Results show a rather small gap between $T^{ub}_{vs}$ and $T^{*}_{vs}$, which suggests that the lower bound $T^{lb}_{vs}$ provides a reliable estimator of $T^{*}_{vs}$. 
They also show the tendency to narrow the gap between $T^{lb}_{vs}$ and $T^{ub}_{vs}$ as $T^{lb}_{vs}$ increases, with the bound ratio $\eta_{T} = T^{ub}_{vs}/T^{lb}_{vs} < 10$ in some of the final 30 cases.

\subsection{Impacting Factors to Performance Map}\label{subsec:bound_map_factors}

\textbf{Inertia and Incidence Angle: }
The inertial flywheel is the main distinction between PIPF and PIP models. Accordingly, it serves as the major factor of interest toward determining landing performance and performance map morphology.

To compute the inertia, we treat the flywheel as a rod with uniformly distributed mass. We believe this approximation is reasonable since many bipedal and quadrupedal robots can be modeled with slim linkages in the sagittal plane (e.g.,~\cite{poulakakis2009spring,nguyen2019optimized,pratt2006capture}). The rod length is proportional to $r_0$ as $l = \eta_{l} r_0$; $\eta_{l}$ is the scaling factor. 
The non-dimensional inertia is
\begin{align}
     \Tilde{I} =
     m l^2 / ( 12 m r^2_0 ) = 
     \eta^2_{l} / 12 \enspace.
     \label{equ:flywheel_inertia}
\end{align}

Picking $\eta_{l} = \{ 0.7, 1, 1.2 \}$ 
produces $\Tilde{I} = \{ 0.04, 0.08, 0.12 \}$ (results rounded to two decimal places). The case $\Tilde{I} = 0.04$ has been studied in Sections~\ref{subsec:prelim_test} and~\ref{subsec:bound_map}. In this section, we continue the search on all three non-dimensional inertia values under different $\Tilde{v}_{z0}$. Since the body incidence angle $\theta_0$ is defined by $\Tilde{v}_{z0}$ and $\Tilde{v}_{x0}$, the search also indicates the effect comparison between the inertia and the incidence angle.

The right four columns of Fig.~\ref{fig:bound_map_collect} summarize the resulting performance maps for each combination of $\Tilde{I}$ and $\Tilde{v}_{z0}$. 
All plots indicate that the linear boundary approximation is preserved under various $\Tilde{I}$, $\Tilde{v}_{z0}$ and $\Tilde{\alpha}_0$. 
The fitted boundary line is described with the coefficients $m$ the slope and $b$ the intercept, as quantified for each plot. 
Panels (a)--(i) show that along the positive $\Tilde{I}$ axis, there is a noticeable reduction over both the slope level $\abs{m}$ and the intercept $b$. 
This observation suggests that the boundary flattens and  the landing success rate decreases when $\Tilde{I}$ increases. 
Larger $\Tilde{I}$ with smaller $b$ also implies the undermining landing capability to handle large initial angular velocity.
On the other hand, no clear boundary changing pattern could be summarized along the $\Tilde{v}_{z0}$ axis. For example, panels (a)--(c) show that both $\abs{m}$ and $b$ have a rising tendency, which cannot be observed in panels (d)--(f).

These observations suggest that the inertia is more dominant than body incidence angle in determining landing capability. Thus, it should be given priority to optimize for when building a robot or adjusting a robot's body morphology for improved landing performance.

\textbf{Initial Attacking Angle: }
Based on~\eqref{equ:capture_point} which decreases monotonically over $\alpha_0 \in (0, 90^{\circ})$, a smaller $\alpha_0$ should handle larger $\Tilde{v}_{x0,min}$. Nevertheless, panels (j)--(l) in Fig.~\ref{fig:bound_map_collect} only illustrate ambiguous change over boundary coefficients $m$ and $b$. 
This observation suggests that $\alpha_0$ has no obvious effect on the boundary morphology as inertia does.

\section{Conclusions}

The paper contributes a nonlinear optimization-based landing framework for the first stance step stabilization on the PIPF model under considerable rotation and translational motions. 
The proposed method is conducted with limited prior knowledge of nominal trajectories and control horizons.
The resulting performance maps over various initial conditions reveal linear boundaries and the dominant effect of inertia on landing capability. 
Future directions of research include 1) a mathematical approximation of the linear boundary for faster evaluation, and 2) an online landing controller that works near the boundary.

\newpage





\end{document}

%% file: Fig_inputs/fig_input_PIPF_model.tex
\begin{figure}[!t]
\vspace{-6pt}
\centering
\includegraphics[trim={0cm 0cm 0cm 0cm},clip,width=0.72\linewidth]{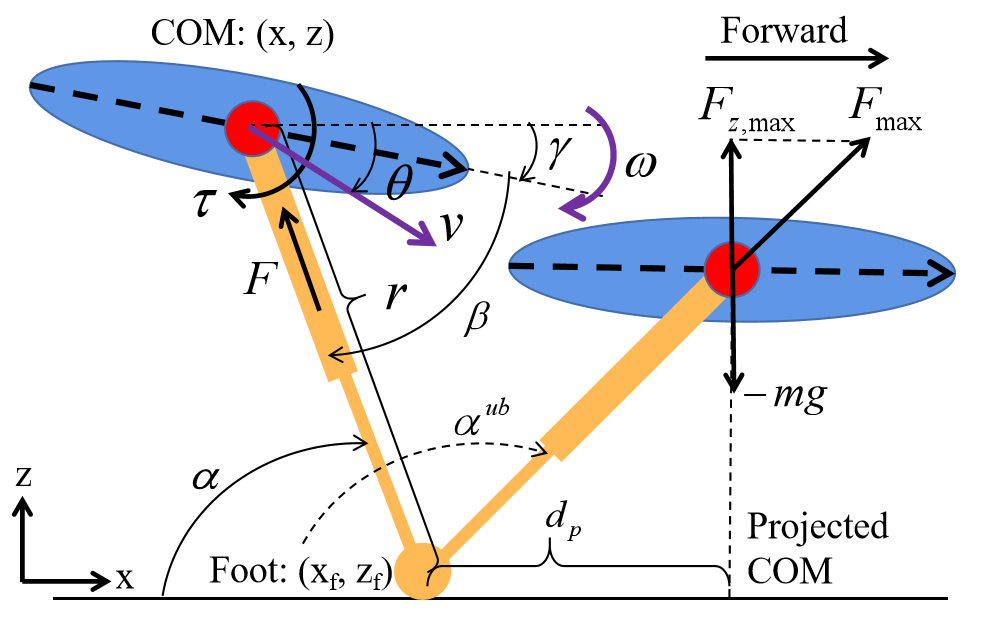}
\vspace{-6pt}
\caption{\textbf{Planer Inverted Pendulum with Flywheel (PIPF) Model.}
PIPF has an inertial flywheel body (blue) with COM (red) and a massless prismatic leg (orange). The body sagittal axis is the dashed arrow.
General coordinates and control inputs are denoted.
The critical stage (right) at $\alpha^{ub}$ signifies that the maximum vertical net force on body COM as $F^{net}_{z,max} = F_{z,max} - mg = 0$. }
\label{fig:model_PIPF}
\vspace{-16pt}
\end{figure}

%% file: Tables/states_summary.tex
\begin{table}[!th]
\vspace{-5pt}
    \caption{Summary of Important Variables} 
    \vspace{-9pt}
    \label{table:variable_summary}
    \begin{center}
    \renewcommand{\arraystretch}{1.5}
    \begin{tabular}{p{2.5cm}>{\centering}p{2.3cm}>{\centering}p{2.3cm}}
        \toprule
        \textbf{Variables} & Dimensional Symbol at $T_k$ & Non-dimensional Symbol at $T_k$ \\ 
        \midrule
        \midrule
        Length or Position
        & $r_k$, $x_k$, $z_k$ 
        & $\Tilde{r}_k$, $\Tilde{x}_k$, $\Tilde{z}_k$ \\ 
        Linear Velocity
        & $v_k$, $v_{xk}$, $v_{zk}$ 
        & $\Tilde{v}_k$, $\Tilde{v}_{xk}$, $\Tilde{v}_{zk}$ \\ 
        Linear Acceleration
        & $\dot{v}_k$, $\dot{v}_{xk}$, $\dot{v}_{zk}$ 
        & $\Tilde{\dot{v}}_k$, $\Tilde{\dot{v}}_{xk}$, $\Tilde{\dot{v}}_{zk}$ \\ 
        \midrule
        Angle
        & $\alpha_k$, $\beta_k$, $\gamma_k$, $\theta_k$ 
        & $\Tilde{\alpha}_k$, $\Tilde{\beta}_k$, $\Tilde{\gamma}_k$, $\Tilde{\theta}_k$ \\ 
        Angular Velocity
        & $\dot{\alpha}_k$, $\dot{\beta}_k$, $\dot{\gamma}_k$,  $\dot{\theta}_k$, $\omega_k$ 
        & $\Tilde{\dot{\alpha}}_k$ $\Tilde{\dot{\beta}}_k$, $\Tilde{\dot{\gamma}}_k$, $\Tilde{\dot{\theta}}_k$, $\Tilde{\omega}_k$ \\ 
        Angular Acceleration
        & $\ddot{\alpha}_k$, $\ddot{\beta}_k$, $\ddot{\gamma}_k$, $\ddot{\theta}_k$ 
        & $\Tilde{\ddot{\alpha}}_k$, $\Tilde{\ddot{\beta}}_k$, $\Tilde{\ddot{\gamma}}_k$, $\Tilde{\ddot{\theta}}_k$ \\ 
        \bottomrule
        
    \end{tabular}
    \end{center}
    \footnotesize{$\quad$ Note that Figure~\ref{fig:workflow_landing_sequence} defines $T_k,k = 0, 1, 2$.}
    \vspace{-12pt}
\end{table}


%% file: Fig_inputs/fig_input_result_state_traj.tex
\begin{figure*}[!t]
\vspace{0pt}
\centering
\includegraphics[trim={3.5cm 6.5cm 3.0cm 6.5cm},clip,width=\linewidth]{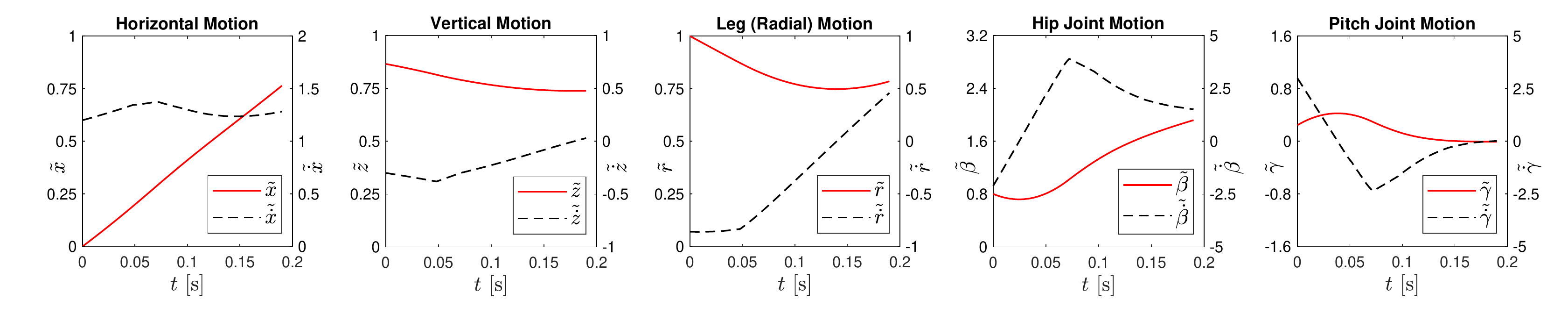}
\vspace{-15pt}
\caption{\textbf{State Trajectories During the First Stance Phase.} Trajectories of non-dimensional states for displacement (red, left axis) and velocity (black, right axis). The setup is consistent with~\eqref{equ:parameter_values} and~\eqref{equ:prelim_test_init_cond}.}
\label{fig:state_traj}
\vspace{-15pt}
\end{figure*}